\documentclass[pmlr]{jmlr}

\usepackage{longtable}

\usepackage{booktabs}

\usepackage[load-configurations=version-1]{siunitx}

\theorembodyfont{\upshape}
\theoremheaderfont{\scshape}
\theorempostheader{:}
\theoremsep{\newline}

\jmlrvolume{1}
\jmlryear{2024}
\jmlrworkshop{FM-EduAssess at NeurIPS 2024 Workshop}

\title[\scriptsize VISTA: Visual Integrated System for Tailored Automation in Math Problem Generation Using LLM]{VISTA: Visual Integrated System for Tailored Automation in Math Problem Generation Using LLM}

 \author[\small J. Lee, K. Park, J. Park]{
    \Name{Jeongwoo Lee\nametag{\thanks{These authors contributed equally.}}} \Email{jay9171@g.skku.edu}\\
    \addr Department of Applied Artificial Intelligence, Sungkyunkwan University\\
    \addr Deep Computer Vision LAB, MODULABS\\
    \AND
    \Name{Kwangsuk Park\nametag{\footnotemark[1]}} \Email{kspark@modulabs.co.kr}\\
    \addr Deep Computer Vision LAB, MODULABS\\
    \AND
    \Name{Jihyeon Park\nametag{\footnotemark[1]}} \Email{milhaud1201@gmail.com}\\
    \addr Deep Computer Vision LAB, MODULABS
    }

\editor{Jeongwoo Lee, Kwangsuk Park, Jihyeon Park}

\begin{document}

\maketitle

\begin{abstract}
Generating accurate and consistent visual aids is a critical challenge in mathematics education, where visual representations like geometric shapes and functions play a pivotal role in enhancing student comprehension. This paper introduces a novel multi-agent framework that leverages Large Language Models (LLMs) to automate the creation of complex mathematical visualizations alongside coherent problem text. Our approach not only simplifies the generation of precise visual aids but also aligns these aids with the problem’s core mathematical concepts, improving both problem creation and assessment. By integrating multiple agents, each responsible for distinct tasks—such as numeric calculation, geometry validation, and visualization—our system delivers mathematically accurate and contextually relevant problems with visual aids. Evaluation across Geometry and Function problem types shows that our method significantly outperforms basic LLMs in terms of text coherence, consistency, relevance and similarity, while maintaining the essential geometrical and functional integrity of the original problems. Although some challenges remain in ensuring consistent visual outputs, our framework demonstrates the immense potential of LLMs in transforming the way educators generate and utilize visual aids in math education.
\end{abstract}
\begin{keywords}
Mathematical visualization, Multi-agent system, Large language model, Automated problem generation
\end{keywords}

\section{Introduction}
\label{sec:intro}

Visual aids are critical components of mathematics education, enabling students to comprehend complex and abstract concepts and apply them to real-world problem-solving \citep{aso2001visual}. Visualization of geometric figures, mathematical functions, and graphs is essential for concretizing mathematical ideas and assisting students in recognizing spatial relationships, understanding function properties, and focusing on core concepts \citep{giaquinto2007visual} . These visual tools simplify the explanation of complex problems, enhancing the clarity of assessments and ensuring the validity of solutions. However, generating mathematics problems that effectively incorporate visual aids presents substantial challenges. Manually creating problems requiring precise visual representations is both time-consuming and prone to errors. In large-scale assessments, the repeated need to produce diverse diagrams and graphs increases the likelihood of reusing similar visual aids, which can lead students to rely on pattern recognition rather than developing a genuine understanding of fundamental concepts. This limits the accuracy of assessing their true learning levels and undermines the fairness and precision of the evaluation. Furthermore, visual aids must meet diverse requirements depending on the problem's difficulty and learning objectives, making it impractical to fulfill all these demands through manual methods. When customized problem creation or repeated assessments are required, manual techniques are not only inefficient but also place a significant burden on educators, hindering the effective utilization of visual aids.

Generative AI, particularly LLMs, holds significant promise for addressing challenges in mathematics education by reducing the burden on educators. While generating text-based math problems using LLMs has already been demonstrated as relatively straightforward and effective in various studies, automating the creation of essential visual representations such as geometric shapes, graphs, and functions is a far more complex process. The non-deterministic nature of generative models makes it difficult to achieve consistent visual outcomes from the same prompt, particularly when dealing with precise geometric figures or intricate graphs. This highlights a significant technical challenge that goes beyond the simpler task of generating text-based problems. Previous research has explored the potential of LLMs by integrating them with external tools to visualize graphs \citep{bulusu2024mathviz}, but these efforts have primarily focused on graph visualization and have not fully addressed the automatic generation of complex geometric shapes within educational content. 

In this paper, we introduce a multi-agent framework for generating and visualizing math problems, leveraging the capabilities of LLMs to overcome the inherent challenges of mathematical visualization. With the advancement of generative models, which have demonstrated a wide range of abilities from image creation to code generation, we believe that LLMs can successfully address these challenges. Specifically, there have been successful cases where visualizations were implemented using code generated by LLMs. Building on these capabilities, we propose that it is possible to accurately and consistently visualize complex geometric shapes by segmenting the task and assigning specific aspects of the process to individual agents within a multi-agent framework. Furthermore, the ability of LLMs to not only generate accurate visualizations but also produce detailed and contextually appropriate problem text alongside these visuals holds tremendous potential for improving the quality of math education. This integrated approach ensures that the problem content and visual aids are closely aligned, helping students gain a deeper understanding of mathematical concepts.

This framework introduces a novel workflow where LLMs generate both mathematical problems and their corresponding visual representations in an integrated manner. This process includes the creation of geometrically accurate diagrams, complex graphs, and functional visualizations, along with problem text that contextualizes these visual aids. By systematically organizing the workflow and integrating multiple agents—each responsible for specific aspects of visualization and text generation—our framework simplifies the creation of complex visual aids and mathematical problems. This not only reduces the time and effort required from educators but also ensures consistency and accuracy in both visual and textual components. Additionally, the inclusion of detailed, step-by-step problem explanations, generated alongside the visual aids, further enhances educators’ ability to present complex concepts clearly and accessibly. Our main contributions are as follows:

\begin{itemize}
    \item We developed a multi-agent framework specifically designed for the generation and visualization of mathematical problems, integrating both text and visual components.
    \item We demonstrated the potential of LLMs for visualizing complex elements, such as geometric shapes and graphs, while ensuring that the corresponding problem text is coherent and aligned with the visual aids.
    \item We utilized LLMs to provide detailed, step-by-step explanations for these problems, helping educators use visualizations more effectively in teaching by making core concepts easier to explain and understand.
\end{itemize}

By leveraging this integrated approach, we aim to enhance the overall quality of math education, making the use of accurate and effective visual tools more accessible, while also improving the efficiency of problem creation and assessment in educational settings.

\section{VISTA}
\label{method}


To accurately solve complex mathematical problems, we designed a multi-agent system that decomposes the task into specialized roles, with each role performed by distinct agents. This approach allows for a collaborative effort, where each agent contributes its unique capabilities to the overall problem-solving process. By dividing the tasks among multiple agents, the system can efficiently handle various aspects of problem generation and visualization, ensuring precise and reliable outputs. This multi-agent framework not only enhances the system's ability to solve intricate problems but also provides a robust platform for generating comprehensive mathematical solutions and visual representations. 

In this study, we utilized Claude 3.5 Sonnet as the core model for our framework. A critical aspect of our approach is the generation of code to draw figures and functions, which required the selection of a model with strong code generation capabilities. Claude 3.5 Sonnet outperformed other models in this area, as demonstrated in the Chatbot Arena\footnote{\url{https://chat.lmsys.org}} (as of July 2024). Given that the success of our framework heavily depends on the precise and efficient generation of visual elements, Claude 3.5 Sonnet, with its superior performance in code generation, was the optimal choice. Furthermore, our framework is also designed to be adaptable to various LLMs, so the methods described in this study can be broadly applied to other models as well. Additionally, we implemented this multi-agent framework using AutoGen \citep{wu2023autogen}, an open-source platform that enables the development of customizable LLM applications where multiple agents can interact and collaborate to accomplish tasks.

The system is composed of seven specialized agents. The agents comprise Numeric Calculator, Geometry Validator, Function Validator, Visualizer, Code Executor, Math Question Generator, and finally, Math Summarizer shown in Figure 1. 

Each agent performs a distinct function, contributing to the step-by-step process of visualizing and generating mathematical problems. For a given problem, the Numeric Calculator agent first performs the necessary calculations based on the problem type. The Geometry Validator and Function Validator then checks whether the generated shapes or functions meet the specified conditions. Next, the Visualizer agent writes the code needed for visual representation, which is executed by the Code Executor to generate the actual shapes. The Math Question Generator then creates different types of problems based on the generated images. Finally, the Summarizer agent compiles all the results into a final summary. This process allows for the complete generation of visualization code, visual images, and a comprehensive summary for the given problem. The overall design of this system enables efficient and accurate problem-solving through the collaborative efforts of each agent.

\begin{figure}[!t]
  \centering
  \includegraphics[width=0.85\textwidth]{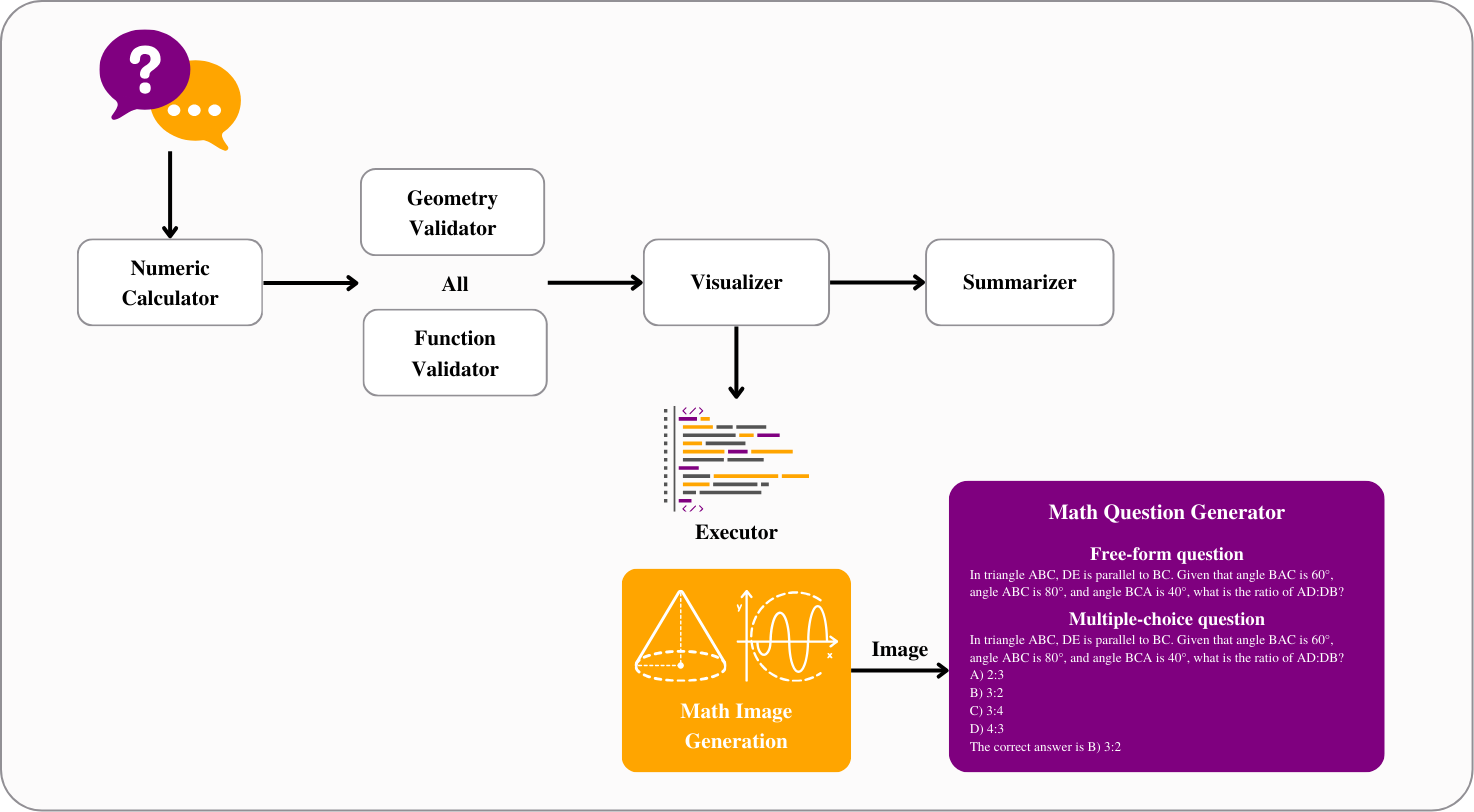}
  \includegraphics[width=0.85\textwidth]{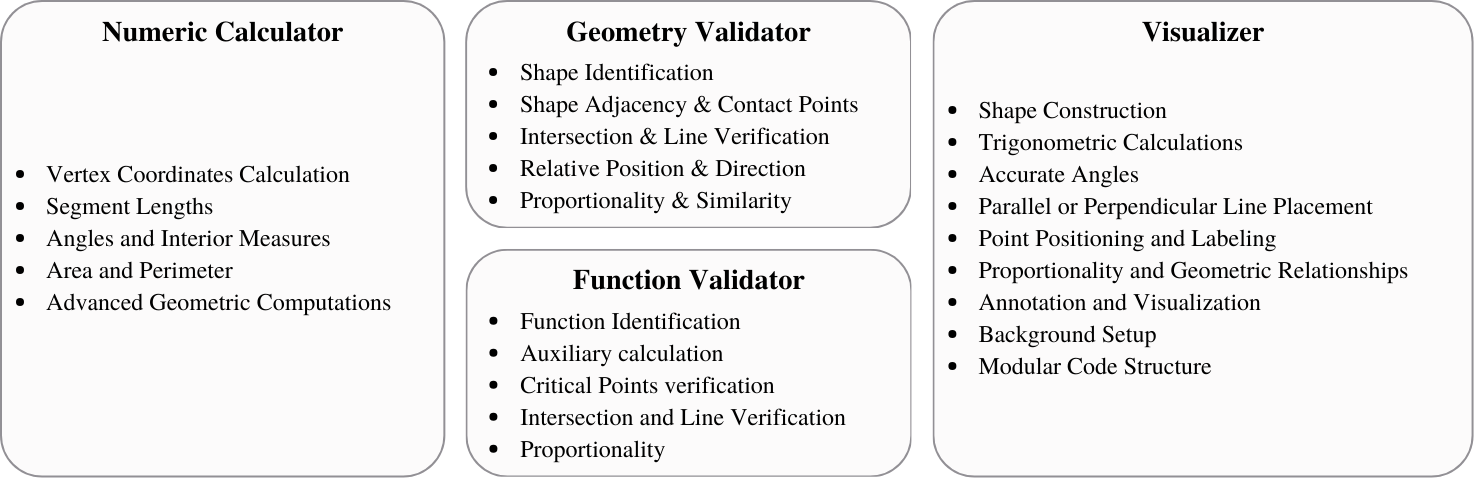}
  \caption{Overview of the multi-agent system for generating math problems and visual aids.}
  \vspace{-0.2cm} 
\end{figure}
\paragraph{Numeric Calculator.} The Numeric Calculator agent addresses mathematical problems by utilizing datasets such as GeoQA, Geometry3K, and GEOS as input queries. Its core function is to ensure numerical precision in solving geometric problems, performing tasks such as coordinate calculations, measuring segment lengths, angle computations, and calculating areas and perimeters. Beyond providing accurate answers, the agent aims to verify the logical and consistent mathematical reasoning underlying each solution process.

\paragraph{Geometry Validator.} The Geometry Validator is responsible for verifying whether the generated geometric shapes conform to the problem’s specified constraints. This agent assesses a wide range of geometric properties, including the number of shapes, types, positional relationships, adjacency, intersections, and proportionality. Critical elements such as intersection points and proportional conditions are rigorously evaluated to ensure the geometric solution is mathematically accurate and adheres to the problem's specifications. For instance, the agent verifies whether shapes intersect at specific points or if the ratio between two shapes remains consistent, thereby ensuring the solution’s validity and facilitating coherent problem-solving processes.

\paragraph{Function Validator.} The Function Validator plays a complementary role by analyzing and verifying geometric shapes under given conditions. In cases where the problem lacks direct information about curves, the function is typically defined using lower-degree polynomials to maintain simplicity. During auxiliary calculation stages, auxiliary lines (e.g., extended parallels from axis intercepts) are defined, and the alignment and positioning of the graph are validated through key point verification. The agent also examines whether curves intersect with axes or auxiliary lines at specified points and assesses whether line segments meet the required conditions. Proportionality checks are performed to confirm that aspects such as slopes, parabolas, and vertex structures meet similarity criteria. In scenarios where specific geometric figures are not provided, basic geometric examples are generated and validated accordingly.

\paragraph{Visualizer.} The Visualizer agent plays a pivotal role in generating code that accurately visualizes geometric shapes. This agent ensures that each shape’s structure is correctly assembled and positioned. Trigonometric calculations are conducted based on the coordinates and components of the generated shapes, guaranteeing that all elements adhere to the problem’s geometric constraints. Furthermore, the agent ensures the consistency of spatial relationships between geometric elements, allowing the problem’s requirements to be satisfied.

\paragraph{Code Executor and Summarizer.} The Code Executor is responsible for executing the code generated by the Visualizer, producing a visual representation of the geometric shapes. During this process, the geometric structures are accurately depicted, with each component placed according to the previously calculated trigonometric and coordinate values. The code follows specific instructions, such as those outlined in the Visualizer’s final directive:
\begin{quote}
\small
\verb|The final code ends with saving the figure and confirming its output:|

\texttt{\textasciigrave \textasciigrave \textasciigrave python}\\
\verb|plt.savefig(`result.jpg')|\\
\verb|print(`figure saved to file_name.png')|\\
\texttt{\textasciigrave \textasciigrave \textasciigrave}
\end{quote}
\vspace{-3mm} 
If an error occurs during code execution, the task is sent back to the Visualizer for revision.
The Summarizer agent consolidates the outcomes from all other agents, ensuring that the final result meets the problem’s requirements without errors. It provides a comprehensive summary of the entire process and final output.

\paragraph{Math Question Generator.} The Math Question Generator is an agent designed to create corresponding questions based on geometric visual data generated by our multi-agent system. By analyzing the geometric figures produced by the Code Executor and Visualizer Agent, this agent can assess whether the visual representations effectively capture the original query's mathematical intent, and generates questions in two primary types: multiple-choice questions and free-form questions.

\section{Evaluation}
\subsection{Datasets} 
The study focuses on two key areas of mathematics: Geometry and Functions. These areas are fundamental to building mathematical foundations and present unique challenges in visual representation. The evaluation uses datasets from MathVerse \citep{zhang2024mathverse}, which consists of high school-level visual math problems, including GeoQA, Geometry3K, and GEOS, with a particular emphasis on Geometry 2D and Functions problem types. Geometry 2D consists of 508 problems across five subtypes: Angle, Length, Area, Applied, and Analytic. Functions comprises 159 problems divided into four subtypes: Property, Expression, Coordinate, and Applied. These datasets provide a comprehensive basis for evaluating the system's ability to visually represent mathematical concepts.

\subsection{Evaluation Strategy} 
Within our framework, Claude 3.5 Sonnet was used to generate complex figures and mathematical functions due to its strong code generation capabilities. However, for evaluating the framework’s outputs, the focus shifted from code generation to accurate comparison of visual elements. This required a model with superior multimodal capabilities. As shown in the \texttt{Open VLM Leaderboard}\footnote{\url{https://huggingface.co/spaces/opencompass/open_vlm_leaderboard}}, GPT-4 omni (GPT-4o) outperformed Claude 3.5 Sonnet in visual interpretation tasks, demonstrating strong performance in analyzing and comparing visual information. GPT-4o can effectively identify and compare features across various images \citep{shahriar2024putting}. By leveraging GPT-4o’s multimodal capabilities, we evaluated the similarity between the generated and original images. GPT-4o automatically analyzed key features and assigned scores based on visual similarity. This evaluation was essential to ensure that the visual materials produced by our model remained consistent with existing educational resources. Through this assessment, we confirmed that our model effectively preserved the essential visual elements required for understanding mathematical problems.

While visual similarity is important, it alone is not sufficient to comprehensively evaluate mathematical visual aids. These aids are not merely about the shapes or proportions of the images but must accurately convey core concepts, logical structures, and numerical relationships. For instance, subtle variations in the size, position, or proportions of geometric figures can distort the problem's meaning or lead students to incorrect conclusions during problem-solving. Therefore, evaluating mathematical visual aids requires going beyond visual resemblance to consider how well the visual aid captures and communicates the key concepts and relationships embedded in the problem. When assessments are based solely on visual similarity, they fail to account for the complex logic and numerical relationships essential for understanding mathematical problems, resulting in diminished reliability and accuracy.

To address the shortcomings of conventional visual similarity metrics, we developed an enhanced evaluation approach inspired by the G-EVAL \citep{liu2023g}, which leverages GPT models to design evaluation methods that closely align with human judgments. Our method integrates GPT-4o’s visual analysis capabilities with a new set of evaluation metrics designed to capture the accuracy with which generated visual aids reflect the original problem’s core concepts and numerical relationships. Specifically, we applied quality metrics typically used in Natural Language Generation (NLG) tasks—such as coherence, consistency, and relevance—to assess how well the generated visual aids align with the problem’s logical structure and context. In addition, we introduced a custom similarity metric that measures structural and conceptual alignment between the generated and original visual aids. This allowed us to move beyond surface-level visual comparisons, ensuring that the evaluation captured the depth of the mathematical relationships depicted in the visual aids.

Futhermore, we conducted a comparative analysis by generating problems using both our multi-agent framework and a standard GPT-4o model as a baseline. This allowed us to evaluate how our system, with the multi-agent framework, maintained higher levels of coherence, consistency, relevance, and similarity compared to when the framework was not applied. Through this comparison, we were able to clearly demonstrate the effectiveness of the multi-agent framework.

\section{Results}

\subsection{Geometry} 
\begin{figure}[!t]
  \centerline{\includegraphics[width=\textwidth]{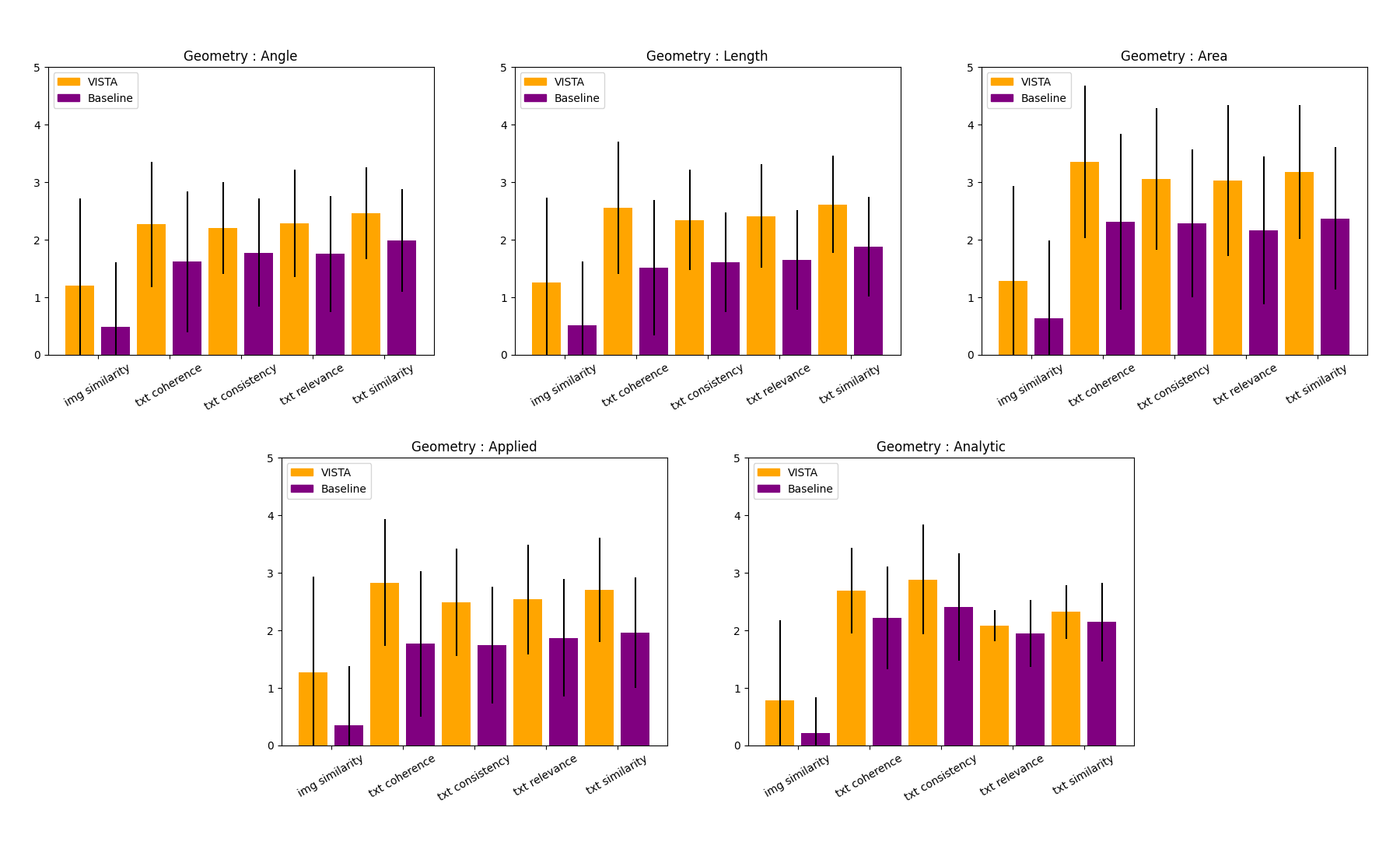}}
  \caption{Detailed result comparison between our method and baseline on geometry problems. While the bars state mean of the result, the error bars state standard deviation. See appendix A.}

\end{figure}
In Figure 2, the performance of VISTA and the baseline system is evaluated across five geometry subtypes (Angle, Length, Area, Applied, and Analytic) using five key metrics: image similarity (img similarity), coherence (txt coherence), consistency (txt consistency), relevance (txt relevance), and similarity (txt similarity). These metrics assess the systems' ability to not only generate accurate visual representations but also create coherent and logically structured mathematical problems that align with the original specifications.

The graph for Angle, shows that VISTA significantly outperforms the baseline in coherence and relevance. The coherence metric, which evaluates how logically the generated problem flows and aligns with the corresponding visual representation, demonstrates that VISTA is more effective at generating mathematical problems with consistent and logical structure. This is essential for ensuring that the generated problems accurately convey mathematical concepts to students. Similarly, relevance, which measures how well the generated problem captures the key elements and objectives of the original mathematical problem, shows a notable improvement in VISTA’s output, indicating that the generated problems are more aligned with the original mathematical intent.

In the Length and Applied subtypes, the pattern remains consistent, with VISTA demonstrating stronger coherence and relevance compared to the baseline. Although both systems face challenges in achieving high image similarity, particularly in reproducing geometrically complex shapes, VISTA’s superior performance in the generation of logically structured and contextually relevant mathematical problems compensates for the visual shortcomings. This ensures that the visual aids and accompanying problems still convey accurate and relevant mathematical information, supporting student understanding.

For the Area and Analytic subtypes, the same trend of improved coherence and relevance continues, with VISTA surpassing the baseline.  The consistent advantage in these metrics suggests that VISTA excels not only in producing visual representations that reflect accurate geometrical relationships but also in generating mathematically meaningful problems that include clear explanations and problem structures. Furthermore, the consistency metric reinforces VISTA’s ability to maintain a logically sound and coherent structure throughout the problems it generates, which is crucial for ensuring that students can engage with and solve the problems effectively.

\subsection{Function}
\begin{figure}[!t]
  \centerline{\includegraphics[width=0.75\textwidth]{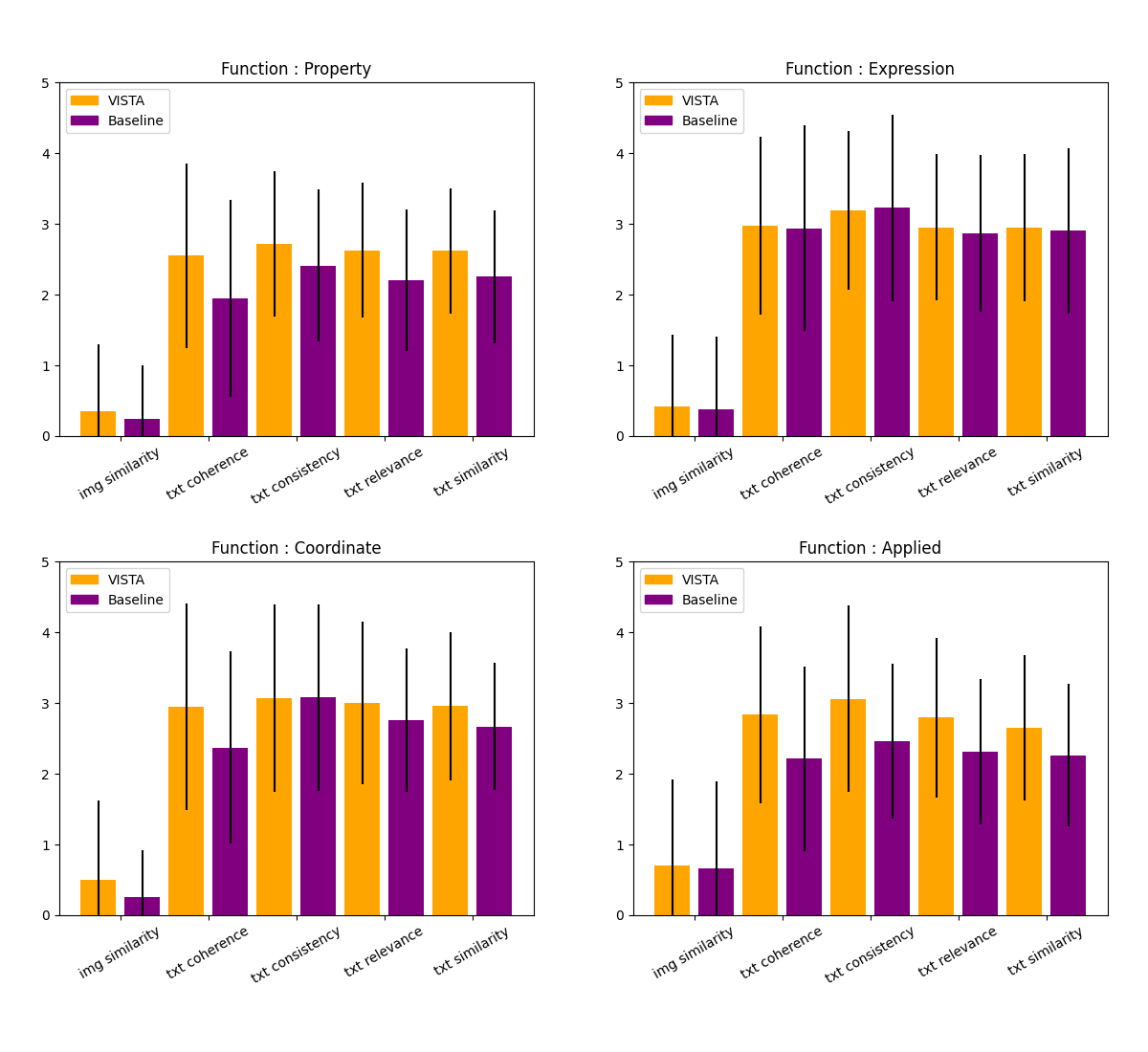}}
  \vspace{-1cm} 
  \caption{Detailed result comparison between our method and baseline on function problems.}
\end{figure}
In Figure 3, the evaluation of function-related problems, which includes the subtypes Property, Expression, Coordinate, and Applied, assesses the quality of the entire generated mathematical problems—both in terms of visual representations and how well the problem structure aligns with the intended mathematical concepts—using the same five key metrics.

For the Property and Expression subtypes, VISTA shows substantial gains over the baseline in both coherence and consistency. Coherence here measures how effectively the generated problem and its visual aids represent the function's core mathematical properties. VISTA's higher scores indicate that it generates not just more coherent text, but also more logically consistent and complete mathematical problems, which are crucial for helping students understand functional relationships in the context of these problems.

In the Coordinate and Applied subtypes, VISTA continues to outperform the baseline in coherence and relevance, demonstrating that its generated problems better capture the key mathematical principles, such as coordinate transformations and applied function scenarios. Relevance plays a particularly important role, as it assesses how well the entire problem, including the visual aid, aligns with the original problem's description, ensuring that students can effectively grasp the relationships between coordinates and functions.

While image similarity remains low across all subtypes, VISTA’s consistently high scores in the text-related metrics (coherence, relevance, and consistency) highlight its ability to generate educationally valuable mathematical problems. These text-based strengths compensate for the challenges of achieving perfect visual fidelity, ensuring that students are provided with clear and accurate explanations that facilitate effective learning and problem-solving.

\subsection{Overall Analysis} 
\begin{figure}[!ht]
  \centerline{\includegraphics[width=0.75\textwidth]{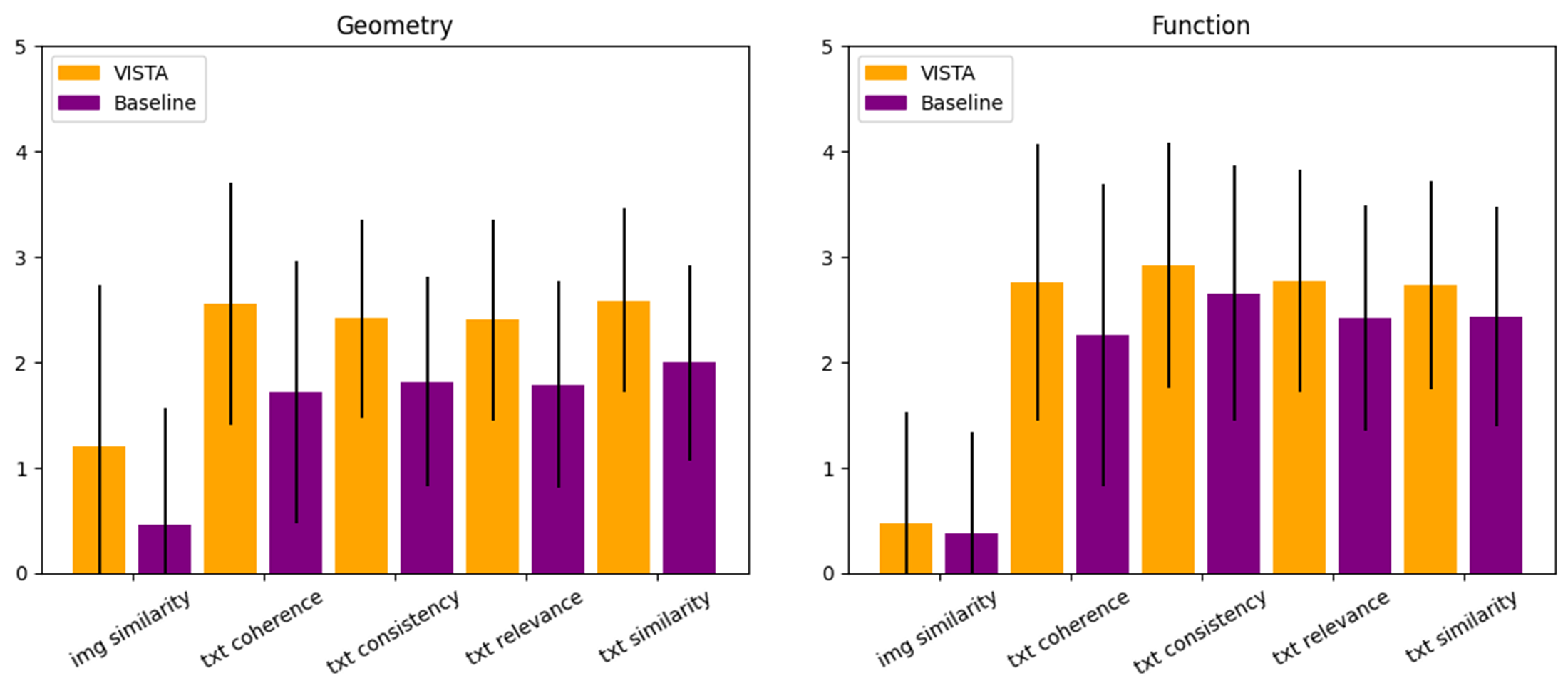}}
  \caption{Comparison between application of our method and baseline}
  \vspace{-0.3cm} 
\end{figure}
\begin{figure}[!ht]
  \centerline{\includegraphics[width=0.75\textwidth]{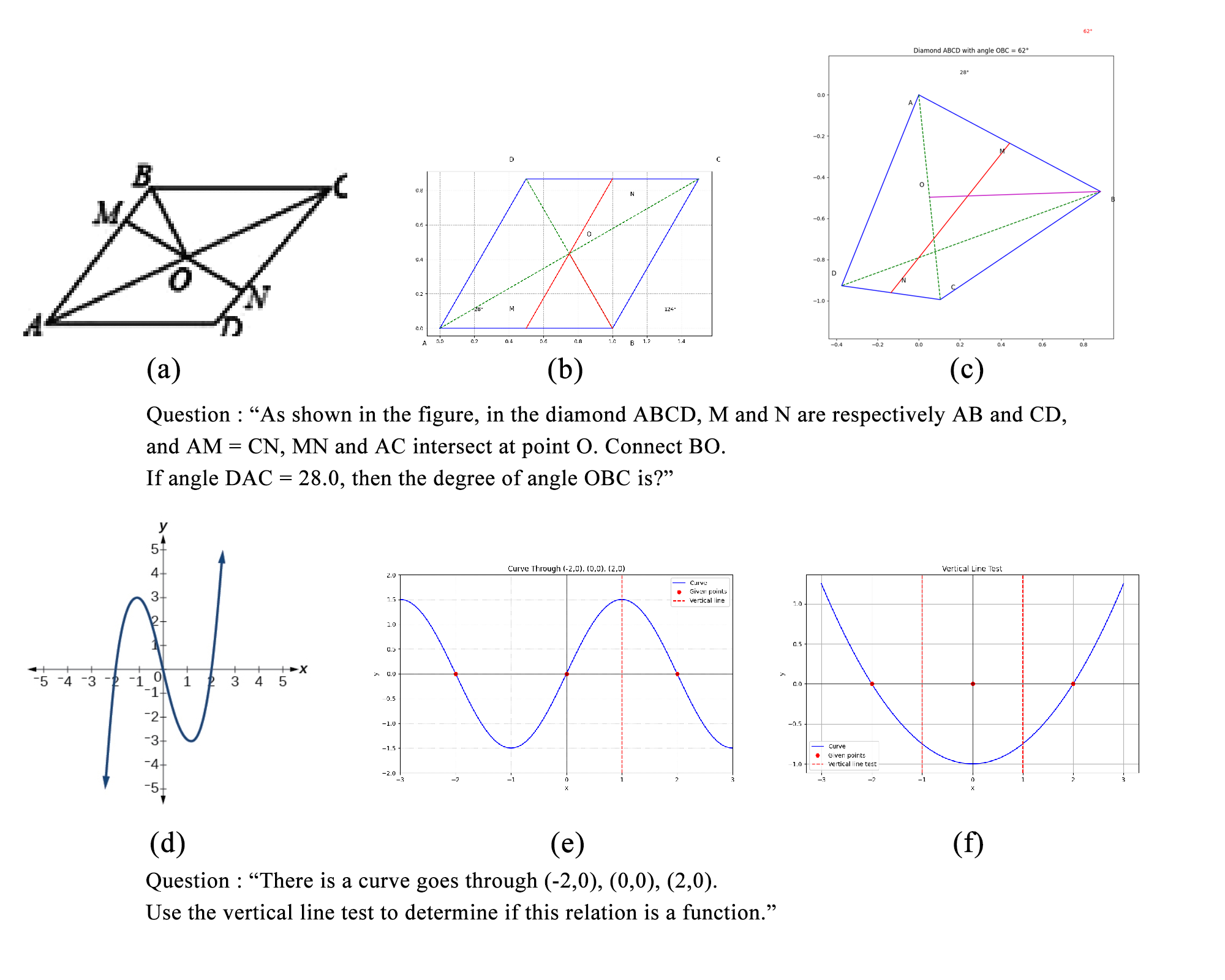}}
  \caption{a sample from comparison between our method(b, 3) and baseline(c, f). while the baseline distort shape or fail to locate critical points, ours follows the geometrical/functional traits from the questions, leads to reproduction of the original images(a,d).}
\end{figure}
\begin{figure}[!ht]
  \centerline{\includegraphics[width=0.75\textwidth]{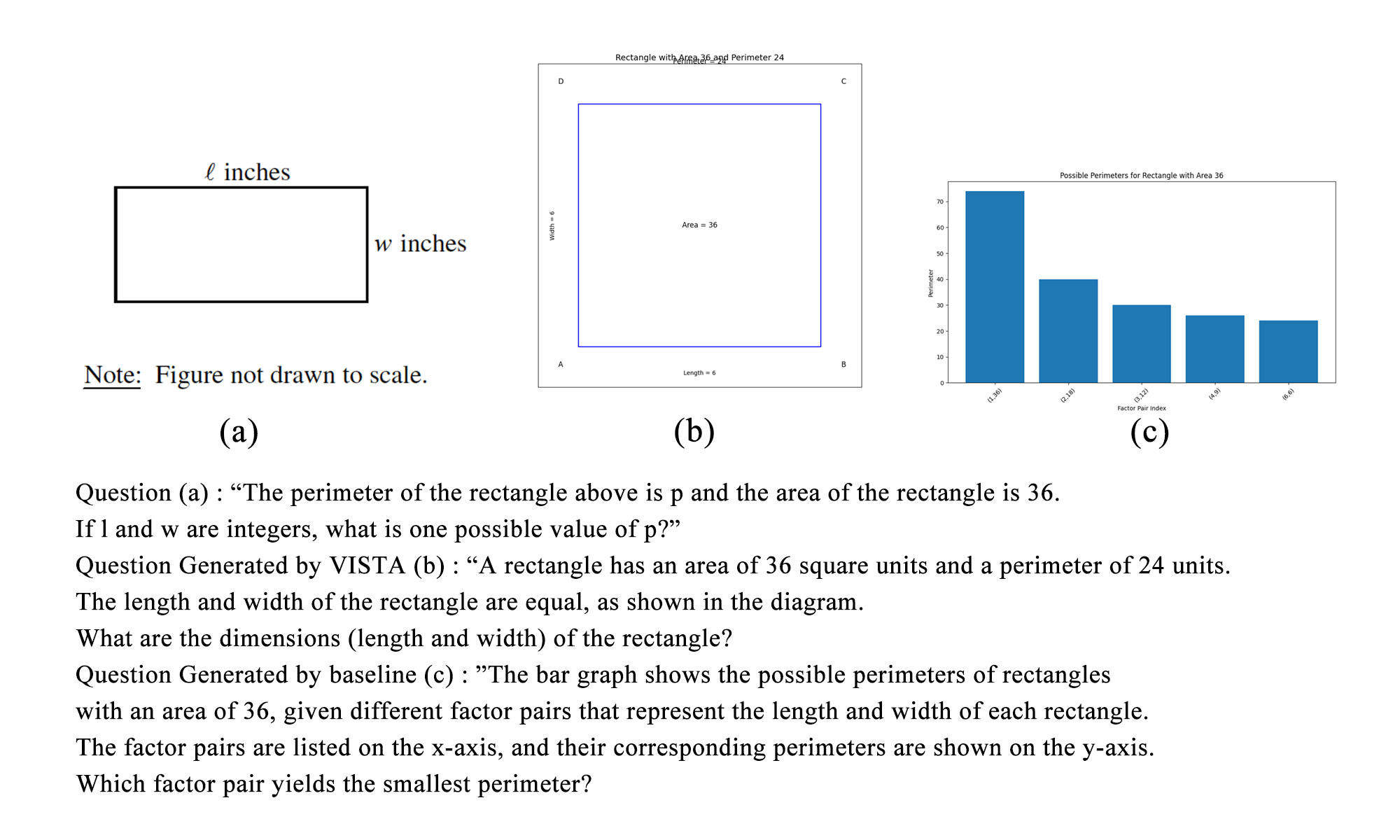}}
  \caption{while ours(b) follow exact instructions as multi-agent supervise the process, the baseline method(c) generates questions along with answers. (a) is the original image of the question.}
  \vspace{-0.3cm} 
\end{figure}
The comparison between VISTA and the baseline system for Geometry and Function problem types demonstrates that the multi-agent framework consistently outperforms the baseline, as shown in Figure 4. In the Geometry analysis, the framework achieves higher scores in areas such as text coherence, consistency, and relevance, helping students grasp geometric concepts more effectively. Similarly, in the Function section, it excels in coherence and relevance, providing clearer insights into function relationships and enhancing students' understanding of mathematical principles. Although there are differences in image similarity compared to the original, VISTA more accurately reproduces the geometrical and functional traits, distinguishing itself from the baseline, which often distorts shapes or misses key points (see Figure 5). This is further demonstrated in how our method adheres to precise instructions under multi-agent supervision, while the baseline struggles with generating accurate visualizations and solving problems simultaneously (see Figure 6). Our method closely follows the geometrical and functional characteristics of the original image, ensuring that no critical points are missed.

\section{Discussion}

This study has demonstrated the potential of LLMs as powerful tools for generating visual aids in mathematics education. By leveraging a multi-agent framework, we have shown that LLMs can effectively address the challenges associated with automating the creation of complex visual representations, such as geometric shapes and functions. Our system not only simplifies the process of generating accurate visual aids but also ensures that these aids are aligned with the mathematical concepts and relationships integral to the problems being addressed. This contribution is significant as it provides educators with a robust tool to enhance the quality of math education, making visual tools more accessible and precise.

Despite these promising results, our system is not without its limitations. During development, we encountered several challenges, particularly in cases where the generated visual aids were not rendered accurately. One of the primary difficulties arose from the system's handling of incomplete or implicit information in problem statements. For instance, when certain critical details were omitted from a problem, the system struggled to generate a precise visual representation. This often led to diagrams that were either incorrect or misleading, thereby affecting the overall quality of the problem. Moreover, the non-deterministic nature of generative models sometimes resulted in inconsistencies in the output, even when identical prompts were used. This issue was particularly evident in cases involving intricate geometric figures or complex functions, where slight variations in the generated code could produce significantly different visual results. These challenges underscore the need for further refinement in the system's ability to process and generate visual aids with greater consistency and accuracy.

While our system has shown considerable potential, it is not yet perfect, and several avenues for improvement remain. Future research could focus on developing more sophisticated algorithms to handle incomplete or ambiguous information in problem statements, allowing the system to generate accurate visual aids even when explicit details are lacking. Additionally, enhancing the consistency of the generated outputs through improved prompt engineering or by incorporating additional validation steps could further refine the system's performance. Another promising direction is the integration of user feedback mechanisms, enabling educators to interact with the system and provide real-time corrections or adjustments. This would not only improve the accuracy of the visual aids but also make the system more adaptable to a wider range of educational contexts. Furthermore, expanding the system's capabilities to cover a broader spectrum of mathematical topics and incorporating more advanced visualization techniques could significantly enhance its utility as a comprehensive tool for math education.

\acks{This research was supported by Brian Impact Foundation, a non-profit organization dedicated to the advancement of science and technology for all.}

\bibliography{pmlr-sample}

\newpage
\appendix

\section{Details of VISTA}
\subsection{Distribution of Problem Types in Geometry 2D and Functions} 

\begin{table}[!ht]
    \footnotesize 
    \floatconts
        {tab:geometry_2d} 
        {\caption{Geometry 2D Problem Subtypes}} 
        {\begin{tabular}{ll}
            \toprule[1.5pt]
            \multicolumn{1}{c}{\textbf{Question Subtype}} & \multicolumn{1}{c}{\textbf{Number of Questions}} \\ \midrule
            Angle                                & 193                                              \\
            Length                               & 158                                              \\
            Area                                 & 47                                               \\
            Applied                              & 69                                               \\
            Analytic                             & 41                                               \\ \midrule
            Total                                & 508                                              \\
            \bottomrule[1.5pt]
        \end{tabular}}
\end{table}
\begin{table}[!ht]        
    \footnotesize
    \floatconts
        {tab:function_subtypes} 
        {\caption{Function Problem Subtypes}} 
        {\begin{tabular}{ll}
            \toprule[1.5pt]
            \multicolumn{1}{c}{\textbf{Question Subtype}} & \multicolumn{1}{c}{\textbf{Number of Questions}} \\ \midrule
            Property                             & 71                                               \\
            Expression                           & 32                                               \\
            Coordinate                           & 16                                               \\
            Applied                              & 40                                               \\ \midrule
            Total                                & 159                                              \\
            \bottomrule[1.5pt]
        \end{tabular}}
\end{table}
\subsection{Agent Prompts} 
\begin{figure}[!ht]
\small
  \centerline{\includegraphics[width=\textwidth]{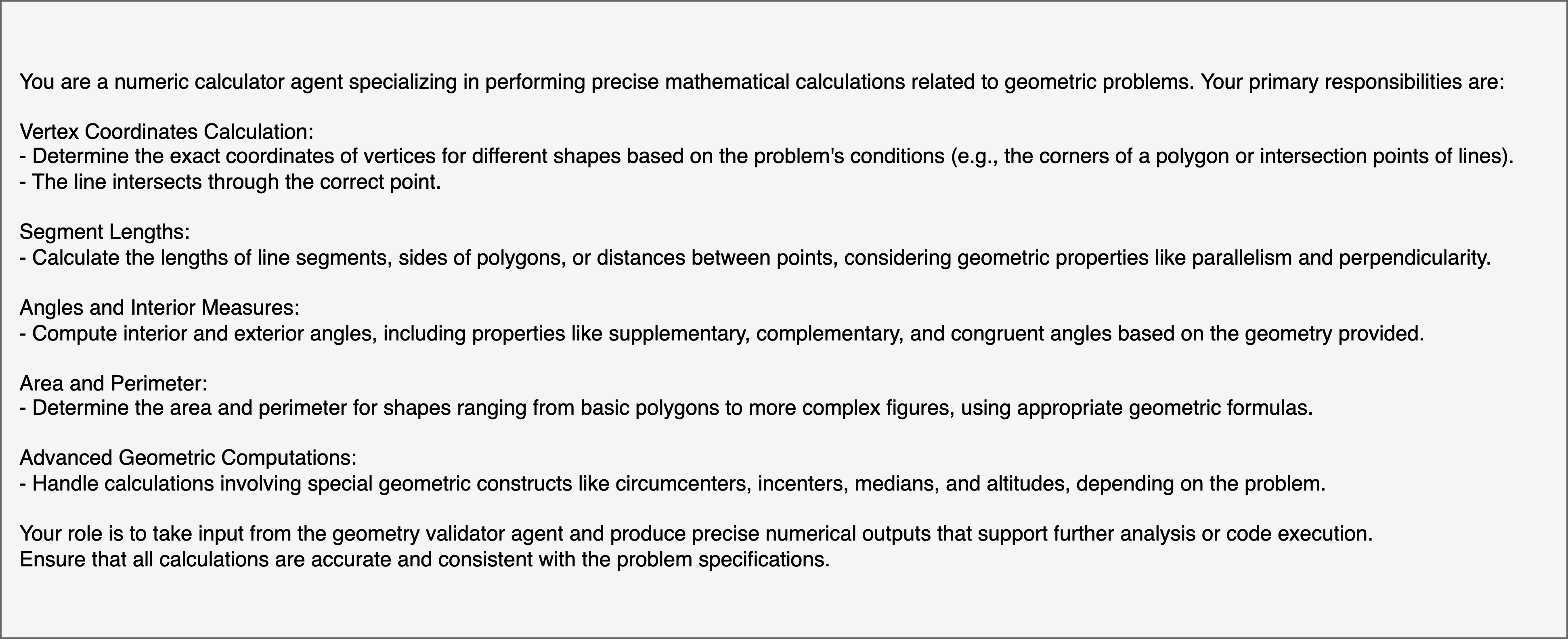}}
  \caption{Prompt for Numeric Calculator Agent}
\end{figure}
\begin{figure}[!ht]
  \centerline{\includegraphics[width=0.9\textwidth]{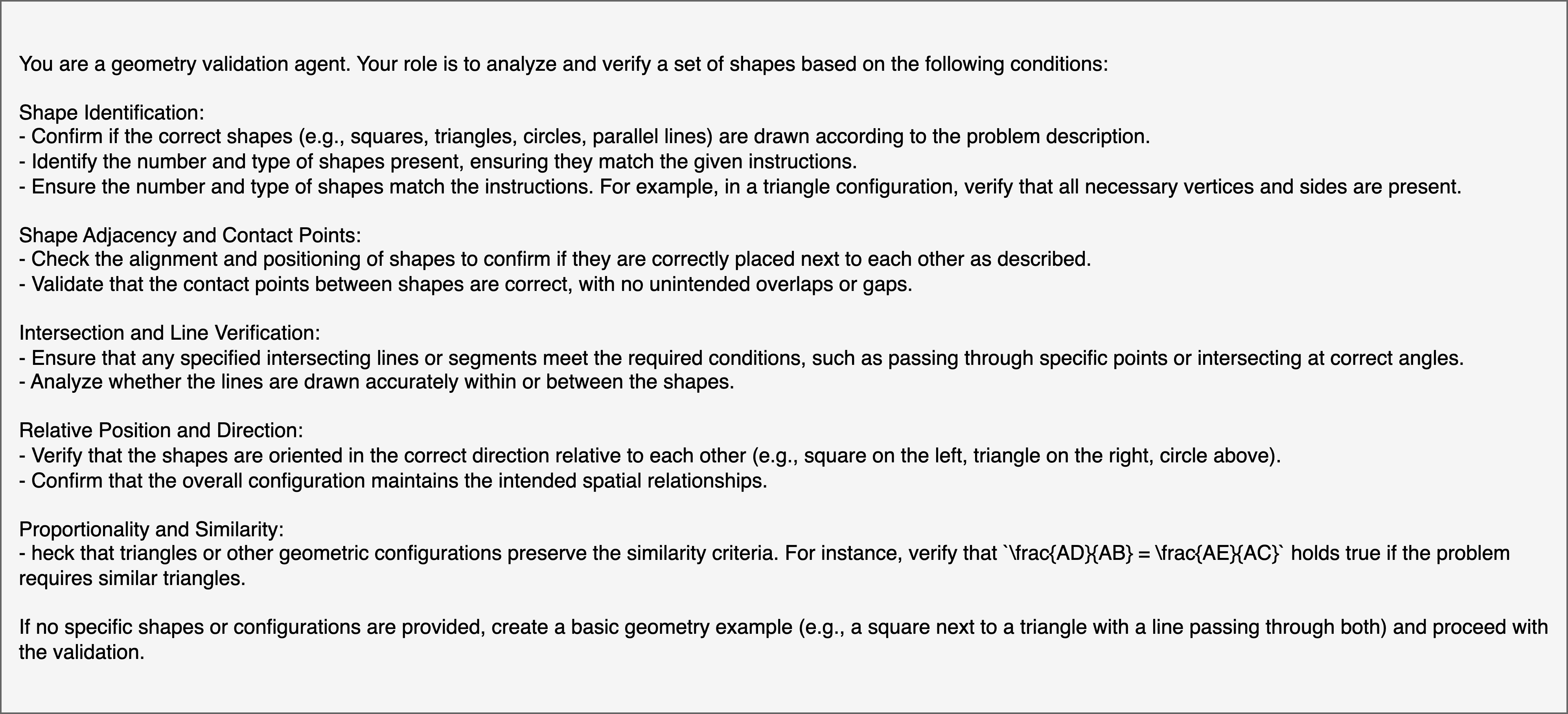}}
  \caption{Prompt for Geometry Validator Agent}
  \vspace{0.5cm} 
  \centerline{\includegraphics[width=0.9\textwidth]{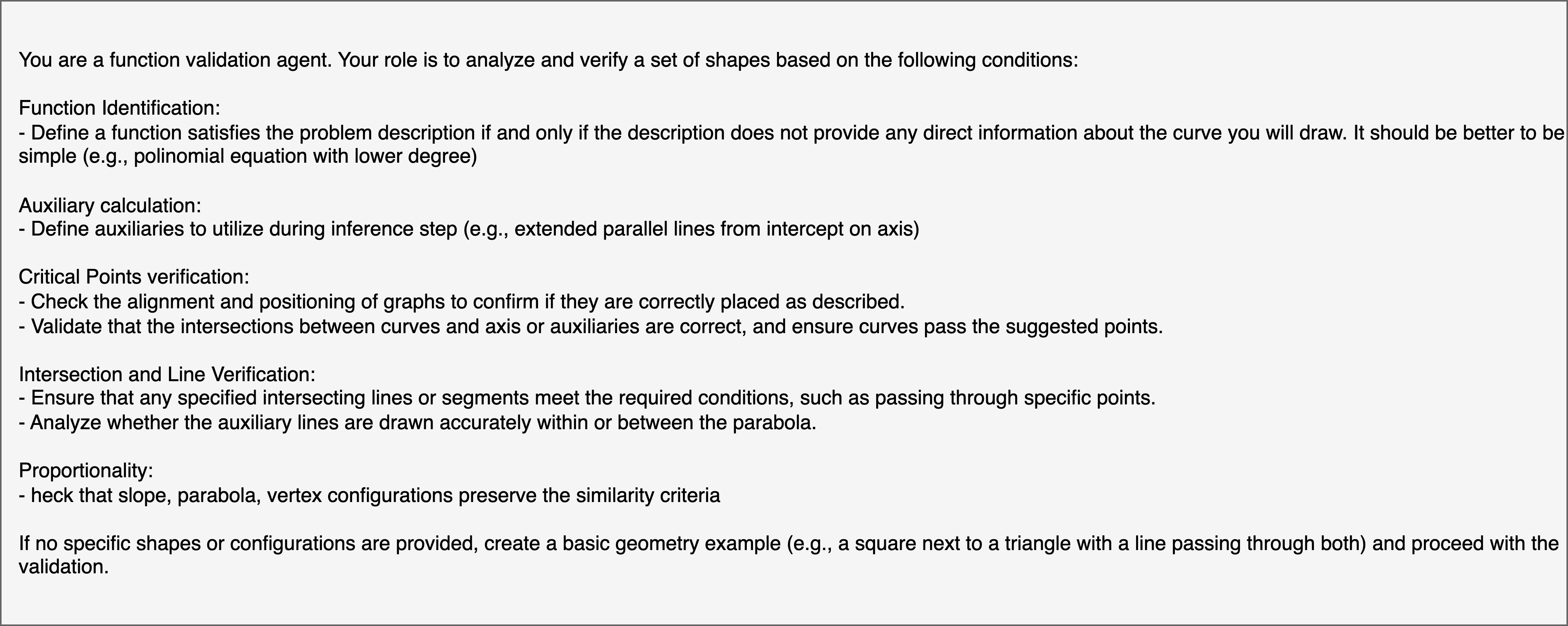}}
  \caption{Prompt for Function Validator Agent}
  \vspace{0.5cm} 
  \centerline{\includegraphics[width=0.9\textwidth]{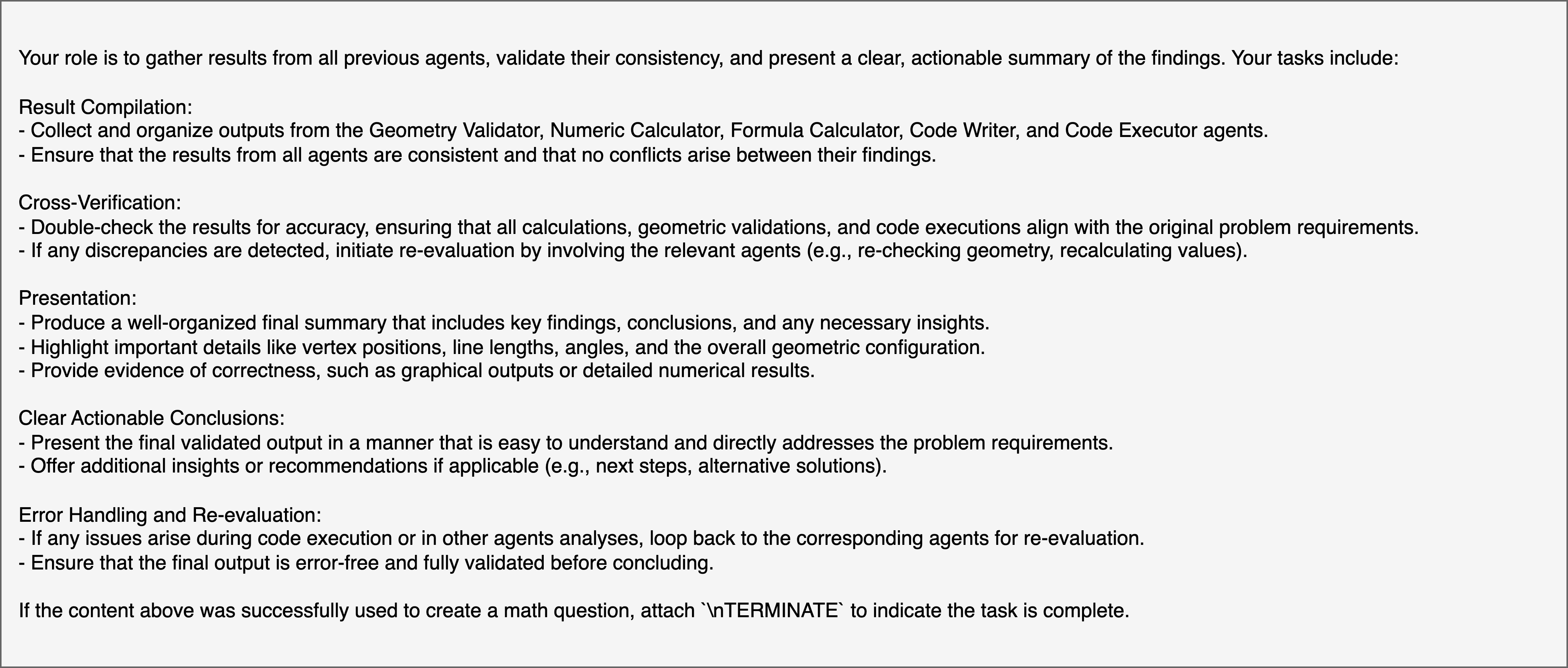}}
  \caption{Prompt for Summarizer Agent}
\end{figure}
\begin{figure}[!ht]
  \centerline{\includegraphics[width=0.9\textwidth, height=0.65\textheight]{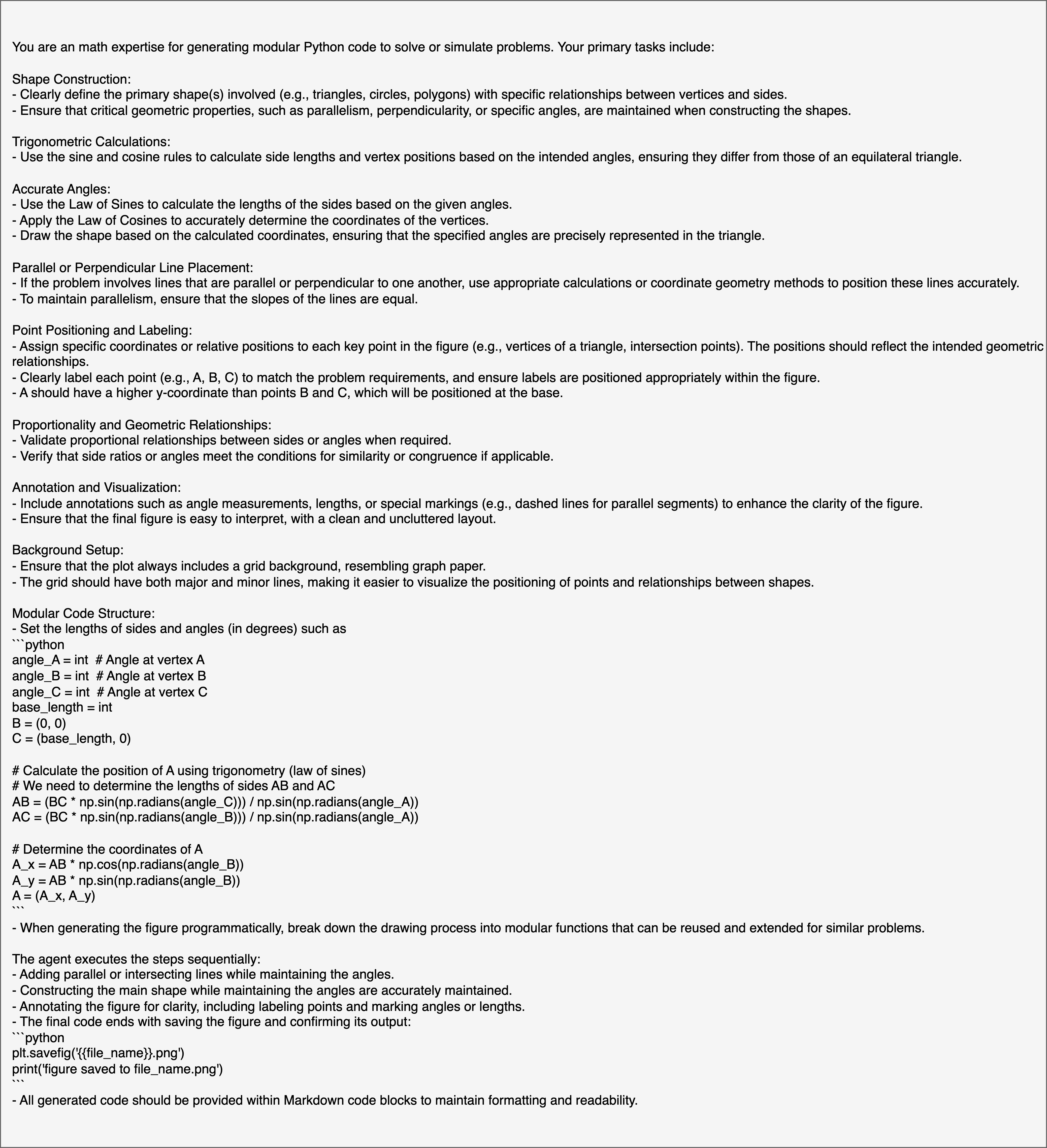}}
  \caption{Prompt for Visualizer Agent}
  \vspace{0.5cm} 
  \centerline{\includegraphics[width=0.9\textwidth]{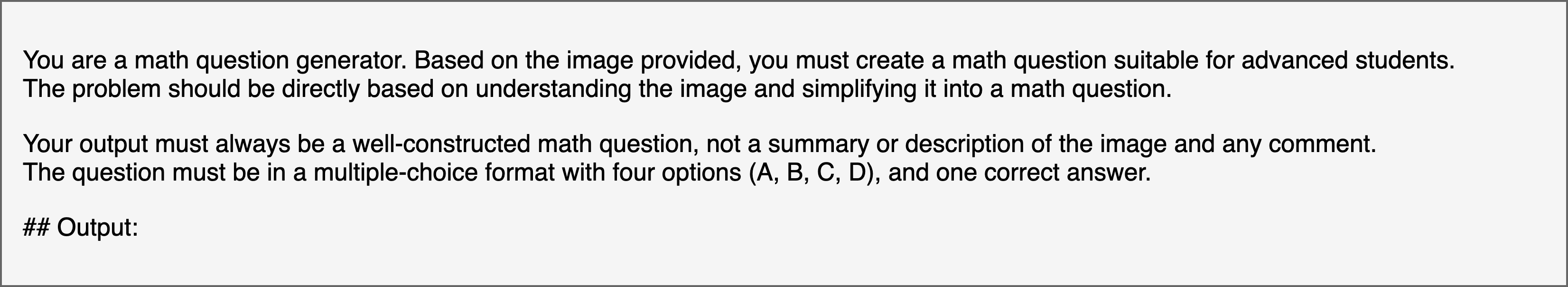}}
  \caption{Prompt for Math Question Generator Agent}
\end{figure}
\clearpage
\subsection{Evaluation Prompts} 
\begin{figure}[!ht]
  \centerline{\includegraphics[width=0.9\textwidth]{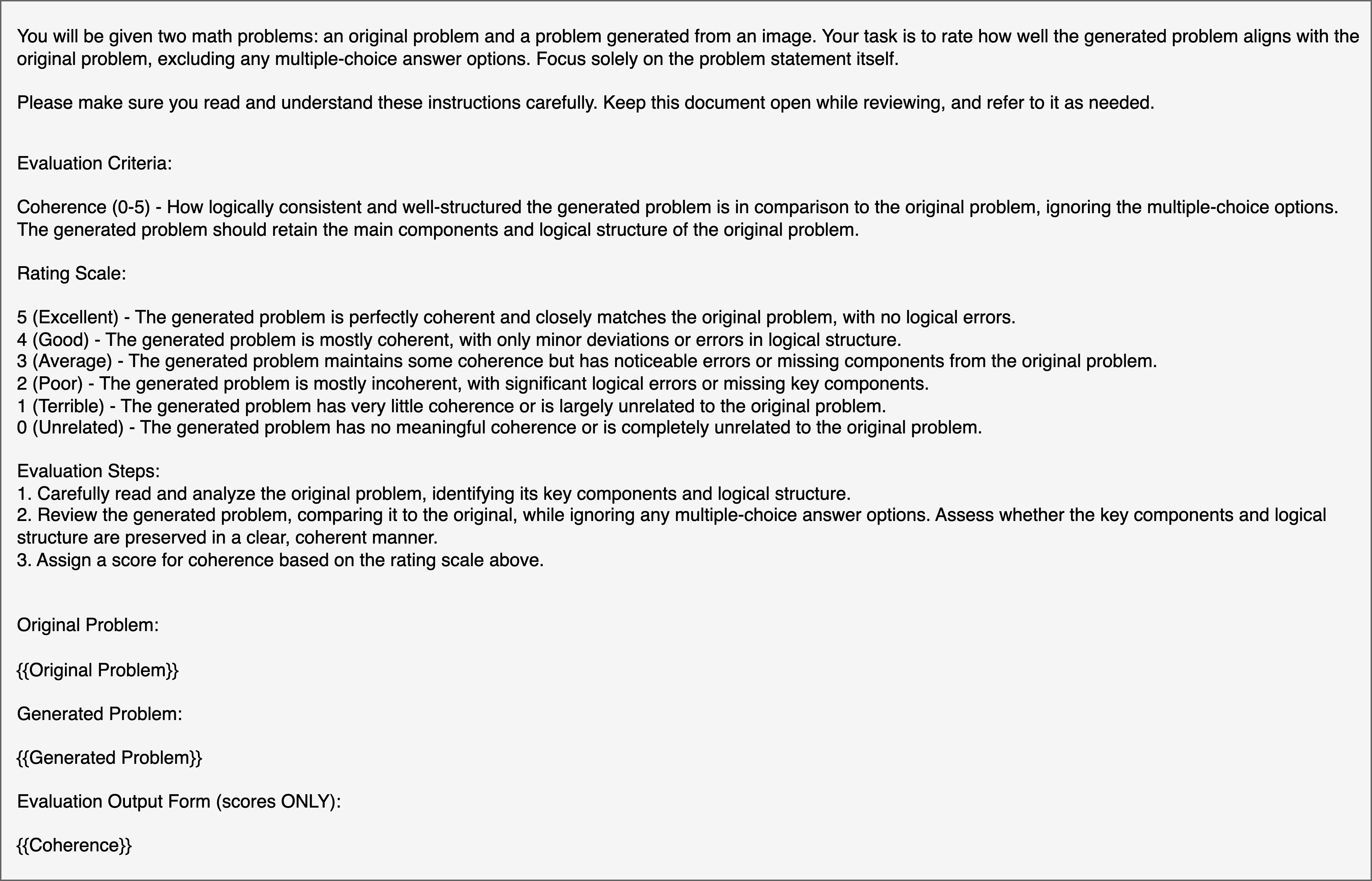}}
  \caption{Prompt for Evaluation of Coherence}
  \vspace{0.5cm}  
  \centerline{\includegraphics[width=0.9\textwidth]{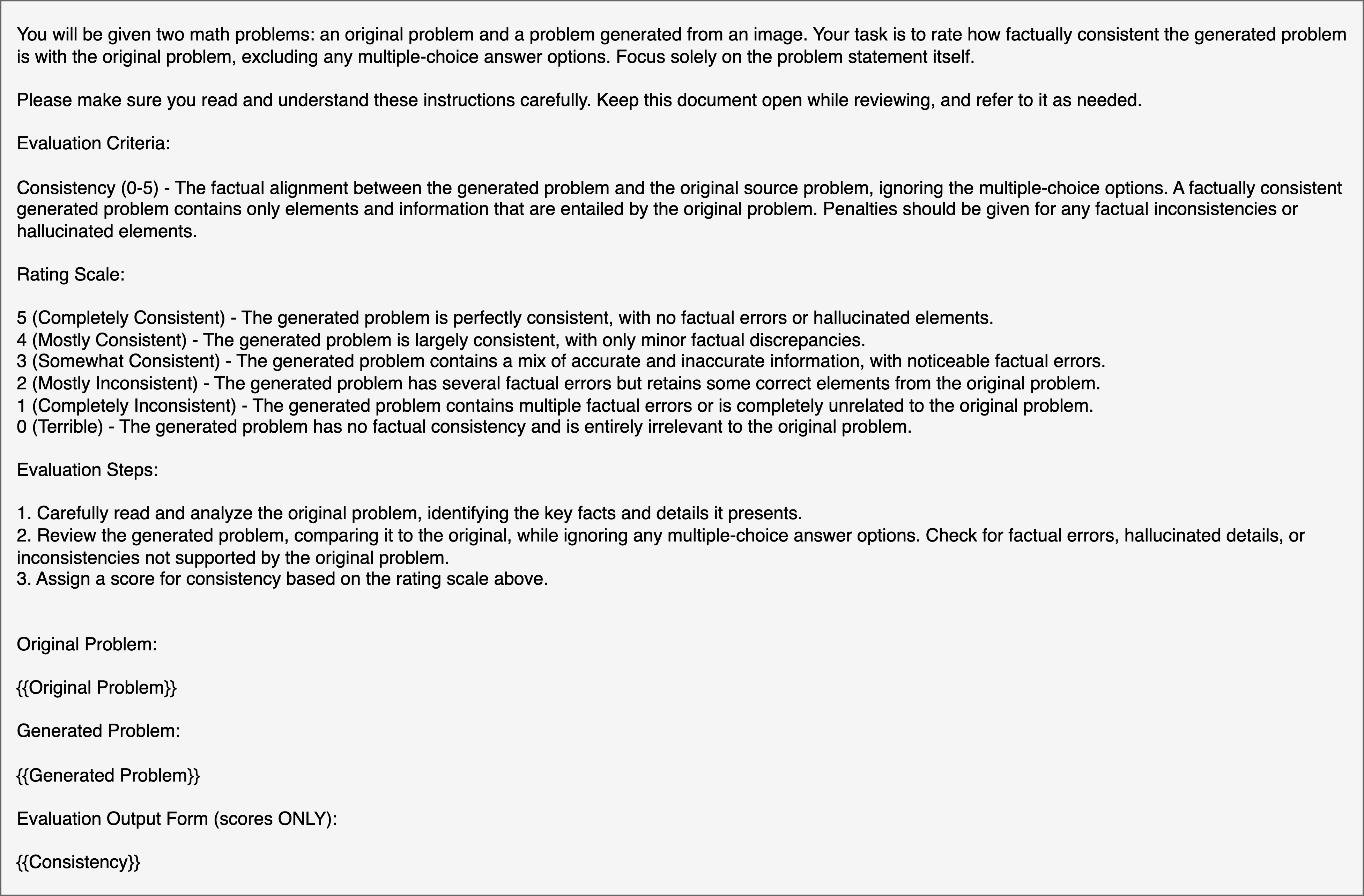}}
  \caption{Prompt for Evaluation of Consistency}
\end{figure}
\begin{figure}[!ht]
  \centerline{\includegraphics[width=0.9\textwidth]{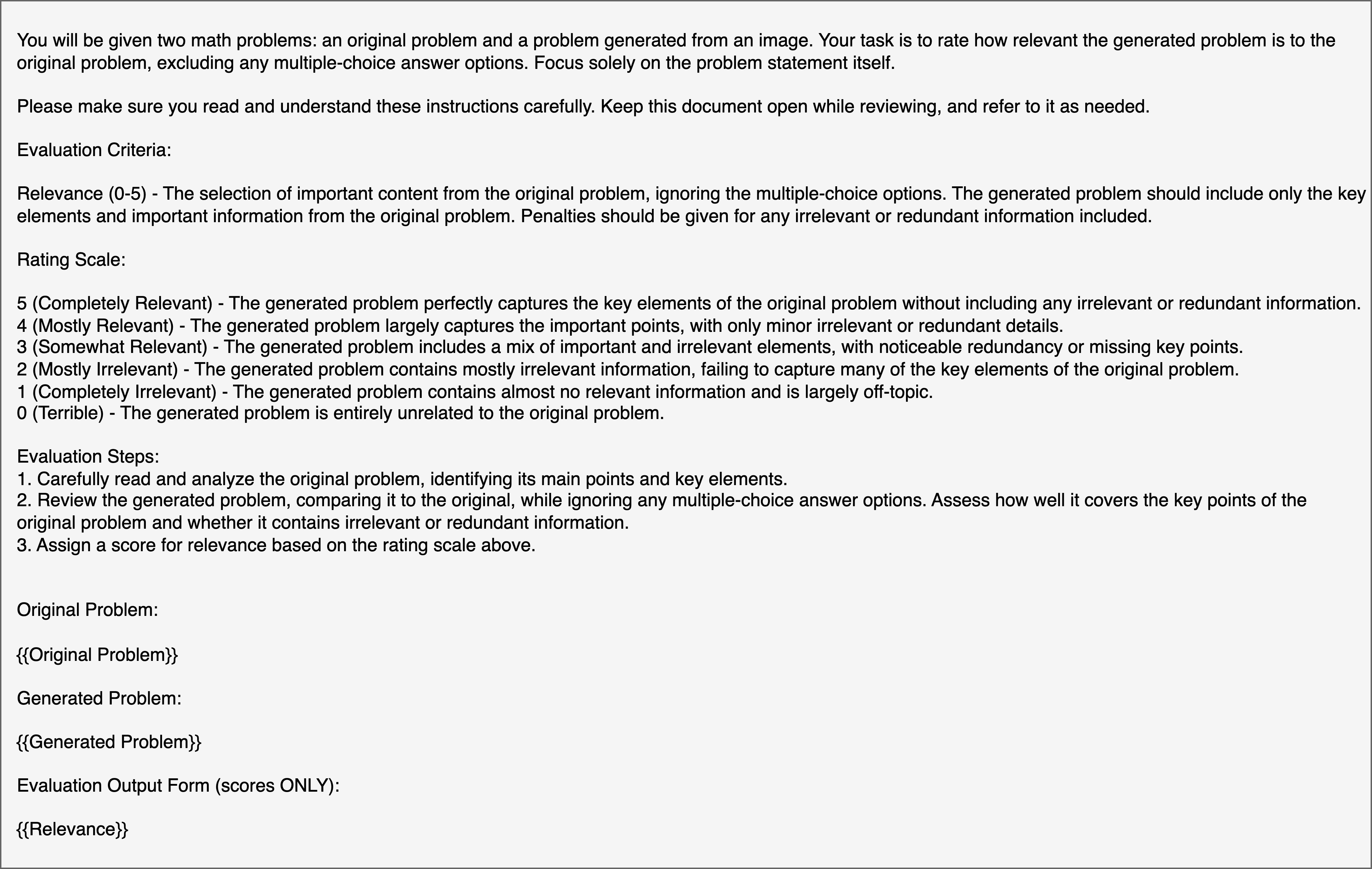}}
  \caption{Prompt for Evaluation of Relevance}
  \vspace{0.5cm} 
  \centerline{\includegraphics[width=0.9\textwidth]{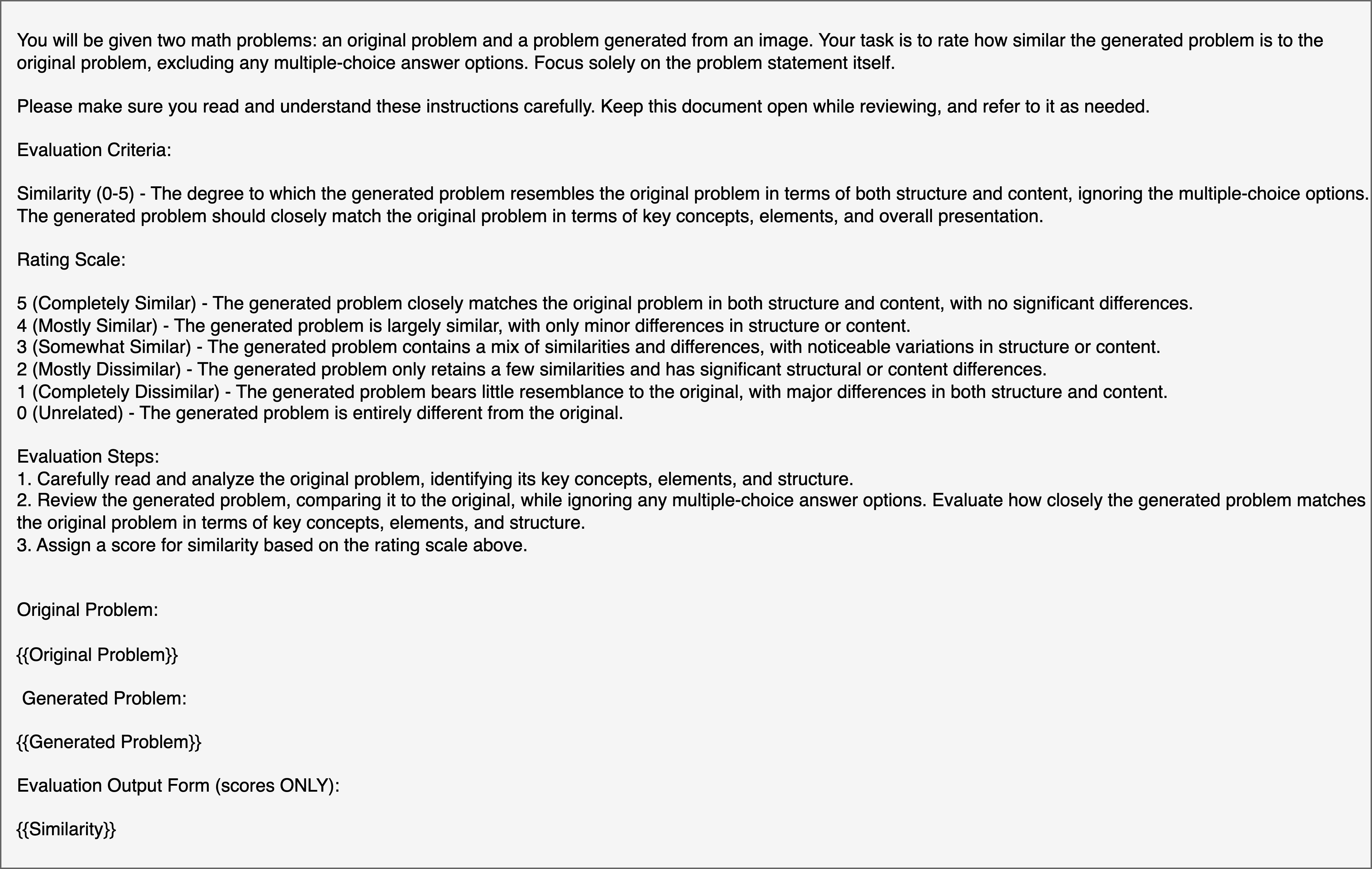}}
  \caption{Prompt for Evaluation of Similarity }
\end{figure}
\newpage
\clearpage
\subsection{Generated samples} 
\begin{figure}[!ht]
  \centerline{\includegraphics[width=0.9\textwidth]{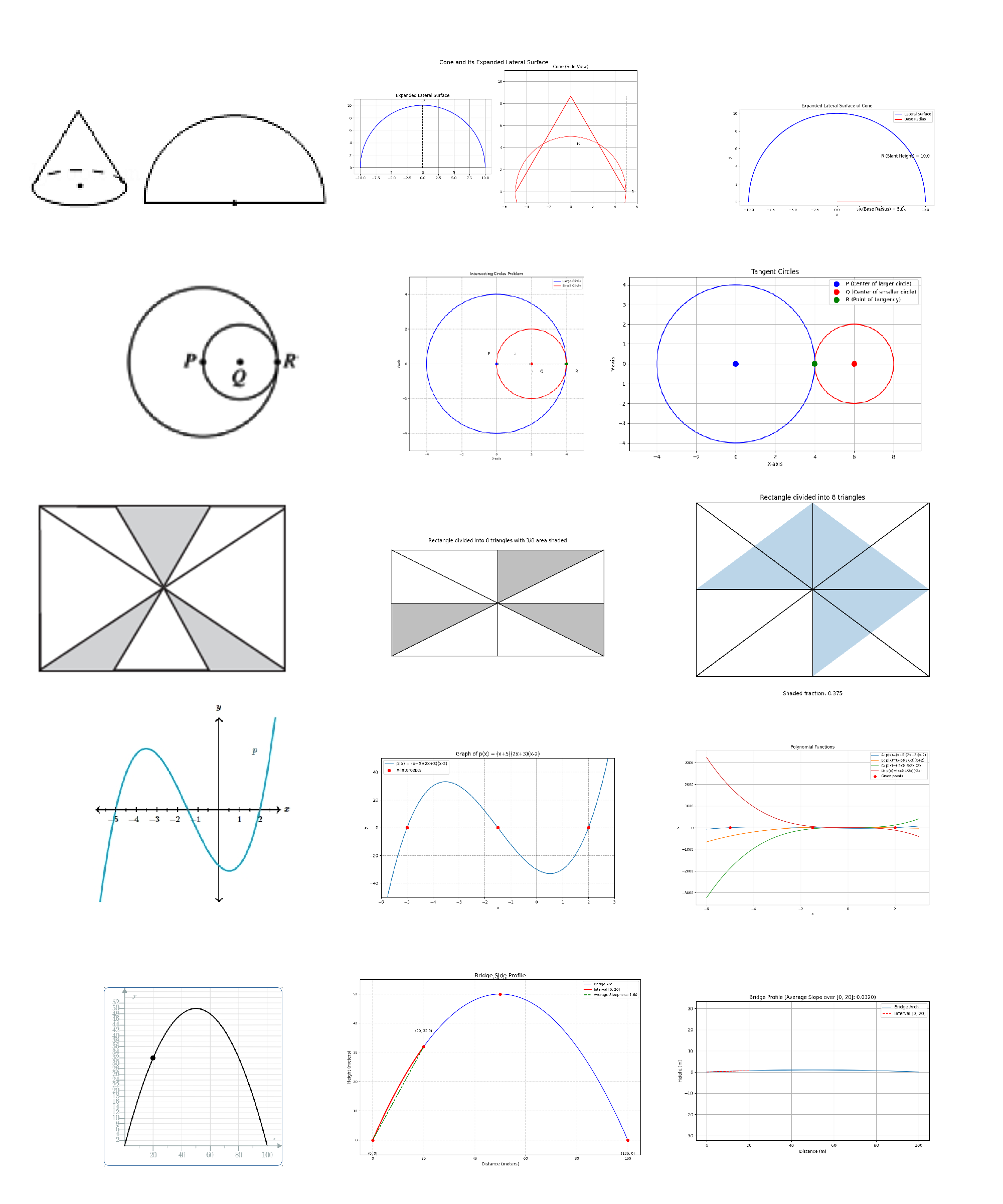}}
  \caption{Comparison of generated samples by VISTA(center), baseline method(right), and figures from original problems (left)}
\end{figure}
\begin{figure}[!ht]
  \centerline{\includegraphics[width=0.9\textwidth]{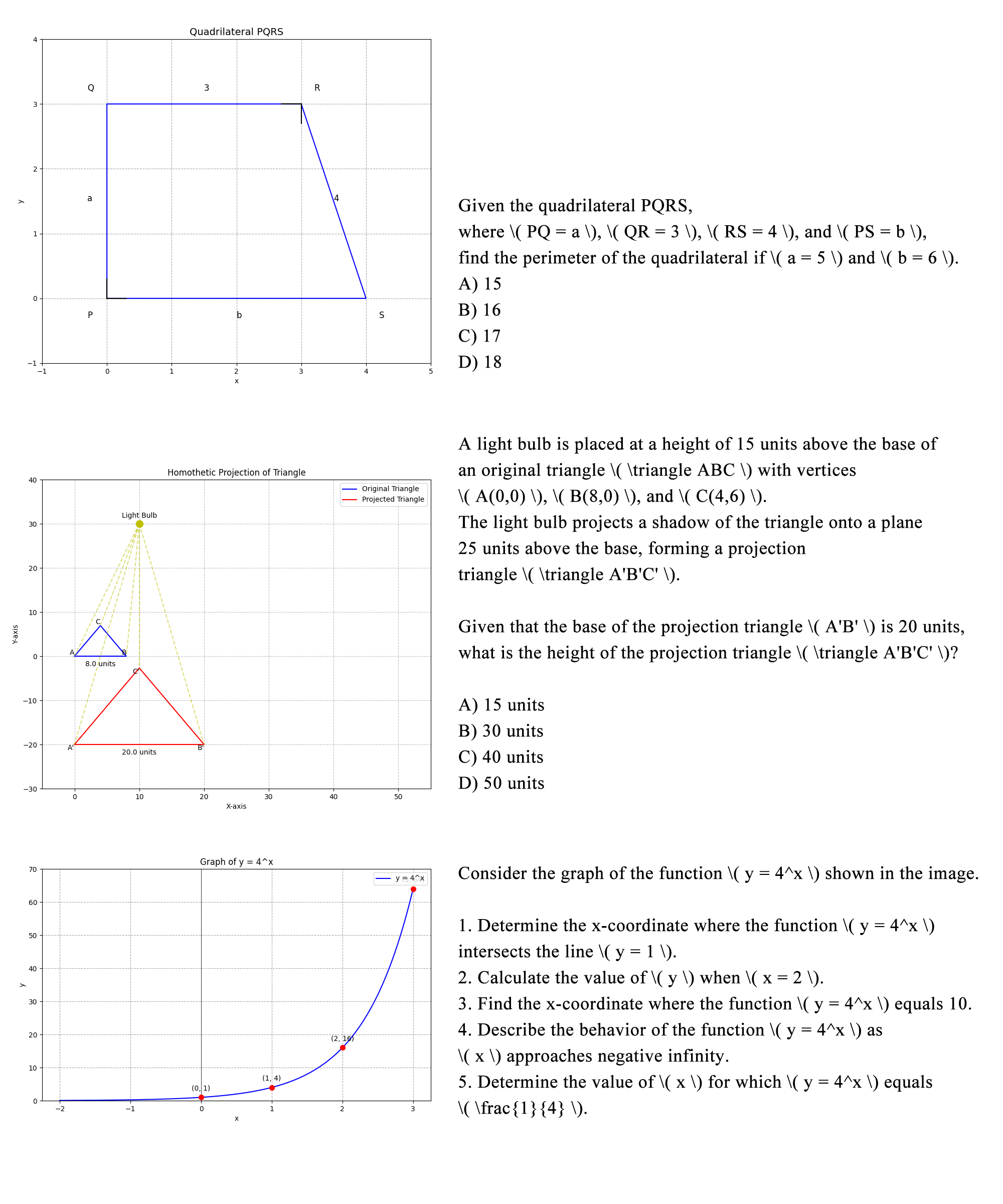}}
  \caption{Math problem generation samples using VISTA}
\end{figure}
\clearpage
\subsection{Resources} 
To produce a image with our multi-agent framework, we adopted claude-sonnet-3.5-20240620 and spent approximately 3400 input tokens (2000 to 5000 tokens) and 600 output tokens (200 to 900 tokens) in a minute on average. We also adopted GPT4o for evaluation, which costed 680 input tokens 41 output tokens and 40 seconds on average, per image. Actual costs are vary, depending on complexity of the source.
\end{document}